\documentclass[]{antgroup}
\PassOptionsToPackage{numbers, compress}{natbib}
\usepackage{antgroup}

\usepackage{graphicx}
\usepackage{tikz}
\usepackage{todonotes}
\usepackage{multirow}
\usepackage{amsmath}
\usepackage{cleveref}
\usepackage{caption}
\usepackage{subcaption}
\usepackage{multicol}
\usepackage{amssymb}
\usepackage{array}
\usepackage{bm}
\usepackage{enumitem}
\usepackage{algorithm}
\usepackage{algpseudocode}
\usepackage{tabularx}
\usepackage{booktabs}
\usepackage{bbm}
\usepackage{makecell}
\usepackage{wrapfig}
\usepackage{colortbl}
\usepackage{threeparttable}
\usepackage{fontawesome5}
\usepackage[normalem]{ulem}
\usepackage{changepage}
\usepackage{xspace}
\usepackage{mdframed}
\usepackage{placeins}
\usepackage{animate}
\usepackage{listings}
\usepackage{pdfrender}
\useunder{\uline}{\ul}{}

\usepackage{amsmath,amsfonts,bm}

\def\eqref#1{equation~\ref{#1}}
\def\1{\bm{1}}

\DeclareMathAlphabet{\mathsfit}{\encodingdefault}{\sfdefault}{m}{sl}
\SetMathAlphabet{\mathsfit}{bold}{\encodingdefault}{\sfdefault}{bx}{n}

\newcommand*\justify{%
  \fontdimen2\font=0.4em% interword space
  \fontdimen3\font=0.2em% interword stretch
  \fontdimen4\font=0.1em% interword shrink
  \fontdimen7\font=0.1em% extra space
  \hyphenchar\font=`\-% allowing hyphenation
}

\renewcommand{\texttt}[1]{%
  \begingroup
  \ttfamily
  \begingroup\lccode`~=`/\lowercase{\endgroup\def~}{/\discretionary{}{}{}}%
  \begingroup\lccode`~=`[\lowercase{\endgroup\def~}{[\discretionary{}{}{}}%
  \begingroup\lccode`~=`.\lowercase{\endgroup\def~}{.\discretionary{}{}{}}%
  \catcode`/=\active\catcode`[=\active\catcode`.=\active
  \justify\scantokens{#1\noexpand}%
  \endgroup
}

\usepackage{amsmath}
\usepackage{amssymb}
\usepackage{amsfonts}                               % blackboard math symbols
\usepackage{amsthm}
\usepackage[mathcal]{eucal}
\usepackage{mathrsfs}
\usepackage{bm}                                     % bm command
\usepackage{blkarray}                               % to support matrix
\usepackage{nicefrac}                               % compact symbols for 1/2, etc.

\usepackage{wrapfig}
\usepackage{graphicx}                               % include pdf figures
\usepackage{caption}
\usepackage{cleveref}
\usepackage{tikz}                                          
\usepackage{circuitikz}
\usetikzlibrary{patterns,snakes}
\usetikzlibrary{positioning,calc,fit,decorations.pathmorphing,shapes.geometric, shapes.gates.logic.US, calc}
\usetikzlibrary{arrows,arrows.meta,decorations.markings,shapes,shapes.arrows}
\usetikzlibrary{decorations,decorations.pathreplacing}
\usetikzlibrary{backgrounds}
\usepackage{filecontents}                           % support to pgfplots
\usepackage{pgfplots}
\usepackage{pgfplotstable}
\usepgfplotslibrary{groupplots}
\usepackage{scalefnt}
\pgfplotsset{compat=newest}
\usepgfplotslibrary{polar}

\usepackage{xcolor}
\definecolor{LightGray}{gray}{0.9}
\definecolor{firstcolor}{HTML}{5E83B5}
\definecolor{secondcolor}{HTML}{2A4B8C}
\definecolor{myBlue}{HTML}{5E83B5}
\definecolor{myBlueBase}{HTML}{2A4B8C}

\newcommand{\lladanew}{LLaDA-UI\xspace}
\newcommand{\projectpage}{\raisebox{-1.5pt}{\faGlobe}\xspace}
\newcommand{\github}{\raisebox{-1.5pt}{\includegraphics[height=1.05em]{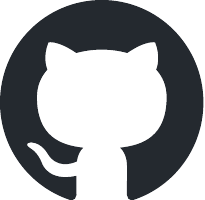}}\xspace}
\newcommand{\hf}{\raisebox{-1.5pt}{\includegraphics[height=1.05em]{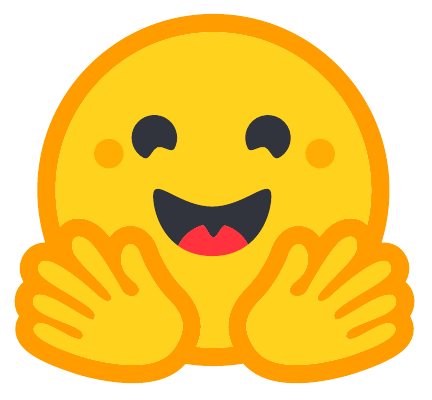}}\xspace}
\newcommand{\bestscore}[1]{{\bfseries\rlap{#1}\kern0.30pt#1}}
\newcommand{\guibestscore}[1]{{\bfseries\rlap{#1}\kern0.55pt#1}}

\title{\lladanew: Bringing Block-wise Diffusion to Vision-Language GUI Agents}
\author{\centering
Zhangxuan Gu$^{2*}$,
Haoxing Chen$^{1*}$,
Qi Qin$^{1*}$,
Yi Xin$^{1*}$,
Kai Gan$^1$,
Lin Liu$^1$,
Long Cui$^1$,
Xiaomei Wang$^1$, \\
Beitong Zhou$^2$,
Yunzhu Zhang$^2$,
Zhengwen Zeng$^2$,
Changlong Gao$^2$,
Weizhi Chen$^2$,
Rongchao Zhang$^2$,
Haoyuan Wu$^1$,
Shuheng Shen$^2$,
Changhua Meng$^2$, 
Weiqiang Wang$^{2\dag}$,
Jianguo Li$^{1\dag}$,
Zhenzhong Lan$^{1,3\dag}$
}

\affiliation{$^1$AGI Research Center, Inclusion AI, $^2$Venus Team, $^3$Westlake University}

\begin{document}
\maketitle
\begin{center}
\footnotesize $*$ indicates co-first authors. $\dag$ indicates technical leads.
\end{center}

\begin{abstract}
Diffusion large language models (dLLMs) achieve high decoding efficiency
through block-parallel, arbitrary-order generation, making them attractive for
latency-sensitive applications. GUI agents represent a natural testbed for this paradigm, as they must repeatedly perceive screen states and emit structured, spatially grounded actions in real time. However, whether dLLMs can be extended into capable multimodal GUI agents while preserving their parallel decoding advantage remains an open question. We present \textbf{LLaDA-UI}, a 16.7B-parameter MoE-based, block-wise diffusion
vision--language GUI agent. LLaDA-UI follows a two-stage training pipeline: general multimodal pre-training aligns a native-resolution vision encoder with the LLaDA2.0-mini-base diffusion language backbone, followed by GUI-Agent supervised
fine-tuning on diverse mobile, desktop, web, and grounding data.  
Across widely adopted grounding benchmarks and navigation benchmarks spanning multiple platforms, LLaDA-UI substantially outperforms Qwen2.5-VL-7B and surpasses Qwen3-VL-8B on four of six reported GUI benchmarks. These results establish block-wise diffusion as a practical generative paradigm for multimodal GUI agents.
\end{abstract}

\vspace{-4mm}
\begin{center}
    \renewcommand{\arraystretch}{1.35}
    \begin{tabular}{rll}
        \projectpage{} & \textbf{Project Page} & \url{https://www.inclusion-ai.org/LLaDA-UI/} \\
        \github{} & \textbf{GitHub} & \url{https://github.com/inclusionAI/LLaDA-UI} \\
        \hf{} & \textbf{HuggingFace} & \url{https://huggingface.co/inclusionAI/LLaDA-UI}
    \end{tabular}
\end{center}

% The radar is intentionally a native PGFPlots graphic. Update benchmark
% values in figs/hero_radar_data.tex; no external drawing program is required.
% Editable data for the overview figure. Each benchmark axis is independently
% normalized so that the best displayed raw score is 100%. The raw maxima are
% shown in the axis labels in doc/hero_figure.tex.
% Axis order: ScreenSpot-V2, ScreenSpot-Pro, AndroidWorld, MobileWorld,
% OSWorld-Verified, WebVoyager.
%
\def\GuiLLaDA{(0,96.49) (60,81.13) (120,92.56) (180,100.00) (240,70.33) (300,100.00) (360,96.49)}
\def\GuiQwenThree{(0,98.94) (60,80.83) (120,82.87) (180,36.72) (240,81.10) (300,79.44) (360,98.94)}
\def\GuiQwenThreeFive{(0,100.00) (60,100.00) (120,100.00) (180,69.92) (240,100.00) (300,81.90) (360,100.00)}
\def\GuiQwenTwo{(0,91.49) (60,41.10) (120,44.12) (180,27.34) (240,7.18) (300,19.33) (360,91.49)}

\begin{figure}[H]
\centering
\begingroup
\definecolor{radarorange}{HTML}{E07A5F}
\definecolor{radargray}{HTML}{8A94A6}
\definecolor{radarteal}{HTML}{3D9A8B}
\pgfplotsset{
  venus radar/.style={
    width=1.95in,
    height=1.95in,
    ymin=0,
    ymax=100,
    ytick={20,40,60,80,100},
    yticklabels={},
    grid=both,
    major grid style={draw=secondcolor!22, line width=0.35pt},
    minor grid style={draw=secondcolor!10, line width=0.25pt},
    axis line style={draw=none},
    tick style={draw=none},
    xticklabel style={font=\fontsize{5.8}{6.5}\selectfont, align=center},
    legend style={
      draw=none,
      fill=none,
      font=\fontsize{6.0}{6.8}\selectfont,
      at={(0.5,-0.18)},
      anchor=north,
      legend columns=1,
      row sep=-1pt,
    },
  },
}

\begin{minipage}[t]{0.40\linewidth}
\centering
{\small\bfseries (a) GUI-agent performance}\\[7pt]
\begin{tikzpicture}
\begin{polaraxis}[
  venus radar,
  xtick={0,60,120,180,240,300},
  xticklabels={SS-V2\\{\scriptsize 94.0},SS-Pro\\{\scriptsize 65.2},AndroidW.\\{\scriptsize 57.8},MobileW.\\{\scriptsize 25.6},OSWorld\\{\scriptsize 41.8},WebVoyager\\{\scriptsize 56.9}},
]
\addplot+[secondcolor, very thick, mark=*, mark size=1.2pt, fill=secondcolor, fill opacity=0.12]
  coordinates {\GuiLLaDA};
\addlegendentry{LLaDA-UI}
\addplot+[radarorange, thick, mark=none, fill=radarorange, fill opacity=0.06]
  coordinates {\GuiQwenThree};
\addlegendentry{Qwen3-VL-8B}
\addplot+[radargray, thick, dashed, mark=none]
  coordinates {\GuiQwenThreeFive};
\addlegendentry{Qwen3.5-9B}
\addplot+[radarteal, thick, densely dotted, mark=square*, mark size=1.1pt]
  coordinates {\GuiQwenTwo};
\addlegendentry{Qwen2.5-VL-7B}
\node[font=\fontsize{4.6}{5.2}\selectfont,text=secondcolor!75,fill=white,inner sep=0.25pt] at (axis cs:90,20) {20\%};
\node[font=\fontsize{4.6}{5.2}\selectfont,text=secondcolor!75,fill=white,inner sep=0.25pt] at (axis cs:90,40) {40\%};
\node[font=\fontsize{4.6}{5.2}\selectfont,text=secondcolor!75,fill=white,inner sep=0.25pt] at (axis cs:90,60) {60\%};
\node[font=\fontsize{4.6}{5.2}\selectfont,text=secondcolor!75,fill=white,inner sep=0.25pt] at (axis cs:90,80) {80\%};
\coordinate (radarScaleTop) at (axis cs:90,100);
\end{polaraxis}
% Draw the outer label after the axis so it sits above every plot and grid.
\node[font=\fontsize{4.6}{5.2}\selectfont,text=secondcolor!75,
      fill=white,fill opacity=0.92,text opacity=1,inner sep=0.5pt,
      anchor=north,yshift=-0.5pt] at (radarScaleTop) {100\%};
\end{tikzpicture}
\end{minipage}\hfill
\begin{minipage}[t]{0.57\linewidth}
\centering
{\small\bfseries (b) Qualitative diffusion decoding}\\[6pt]
\begin{minipage}[t]{0.25\linewidth}
  \vspace{0pt}
  \centering
  \fcolorbox{secondcolor!35}{white}{%
    \includegraphics[height=1.56in]{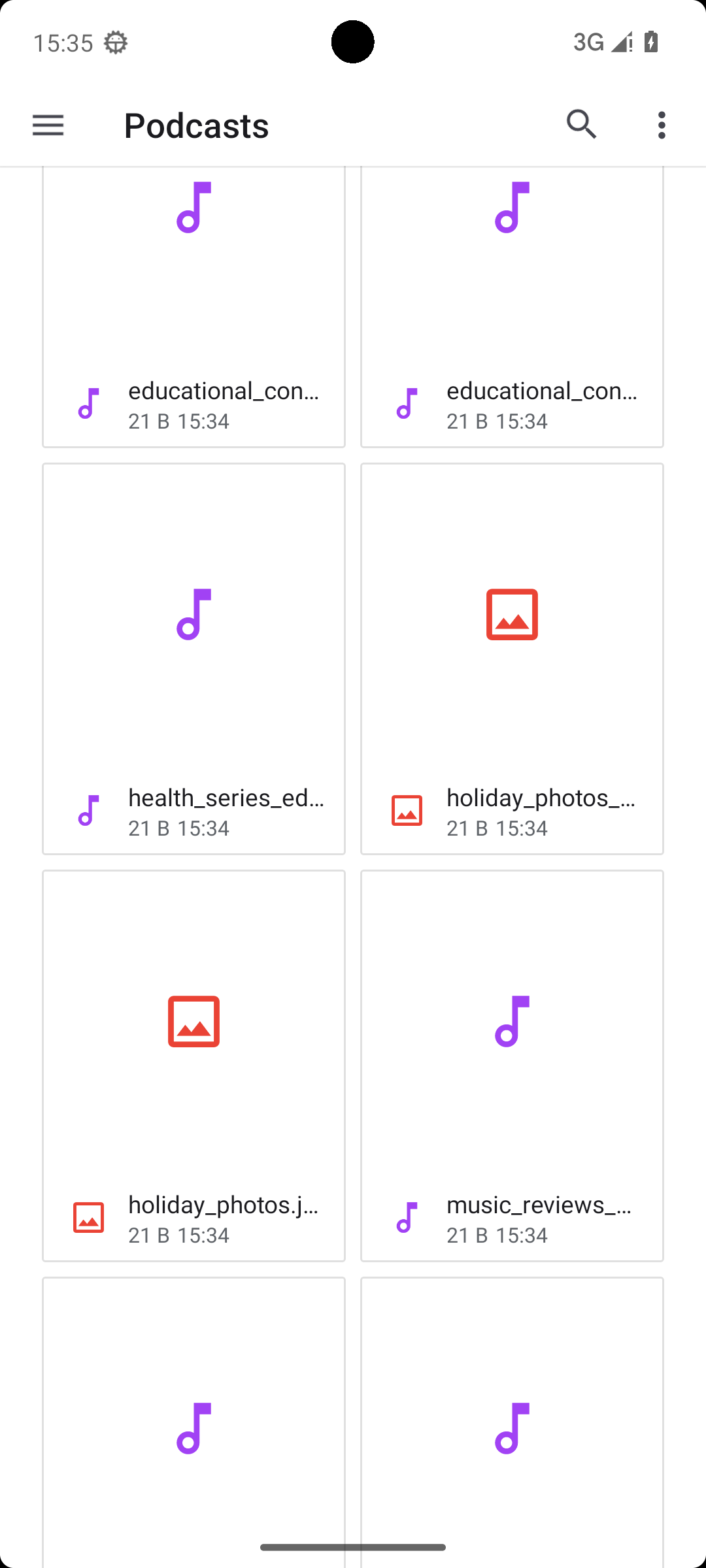}}\\[2pt]
  {\fontsize{6.2}{7}\selectfont Static GUI observation}
\end{minipage}\hfill
\begin{minipage}[t]{0.72\linewidth}
  \vspace{0pt}
  \centering
  \colorbox{secondcolor!7}{\parbox{0.93\linewidth}{%
    \raggedright
    \fontsize{6.2}{7.2}\selectfont
    \textbf{Task.} Move \texttt{holiday\_photos.jpg} from
    \texttt{Podcasts} to \texttt{DCIM} in the same Android storage area.}}\\[6pt]
  \animategraphics[height=1.56in,keepaspectratio,autoplay,loop,poster=last]{3}%
    {figs/diffusion_cases/frames/hero_case10/frame-}{1}{75}\\[3pt]
  {\fontsize{6.2}{7}\selectfont Denoising sequence}
\end{minipage}
\end{minipage}
\endgroup
\caption{Overview of LLaDA-UI. Left: performance of LLaDA-UI and
baselines on GUI-agent benchmarks. Right: a real AndroidWorld prediction case
showing how the complete model output is progressively denoised into the final
response. Best viewed in Adobe Acrobat Reader.}
\label{fig:main_comparison}
\end{figure}

\newpage
\section{Introduction}

Large vision--language models (VLMs)~\citep{hurst2024gpt4o,deepmind2025gemini,team2025qwen3VL} have emerged as a transformative force in artificial intelligence, extending the capabilities of large language models (LLMs) toward unified modeling and joint understanding of visual and textual information. This progress has substantially improved the ability of AI systems to solve multimodal tasks in complex real-world settings. Nevertheless, most existing multimodal large language models still rely on the autoregressive (AR) paradigm. Although highly successful, autoregressive generation is inherently sequential, which limits parallelism and can result in considerable inference latency. Its strictly causal structure can also be restrictive for tasks that benefit from bidirectional context or global semantic refinement.

Masked discrete diffusion large language models (dLLMs)~\citep{LLaDA,LLaDA2} offer a promising alternative by reconstructing complete sequences from masked states, supporting parallel token prediction and bidirectional contextual modeling. Over the past year, dLLMs have gained substantial momentum in text-only language modeling and have begun to support multimodal understanding and generation in a unified model, as demonstrated by LLaDA2.0-Uni~\citep{llada20uni}.  Diffusion vision--language models such as LLaDA-V~\citep{you2025llada} and SDAR-VL~\citep{cheng2025sdar} further demonstrate that diffusion decoding can serve as an alternative to autoregressive multimodal generation.  Recent work has also explored diffusion models for static GUI grounding and single-step coordinate prediction~\citep{kumbhar2026gui}. However, dLLMs remain largely unexplored as vision-language GUI agents that must repeatedly perceive changing screens, produce executable actions, and complete long-horizon tasks across dynamic environments such as MobileWorld~\citep{mobileworld} and OSWorld~\citep{osworld}.

To bridge the gap between multimodal diffusion models and practical application, we present \textbf{LLaDA-UI}, an MoE-based, block-wise diffusion vision--language GUI agent with 16.7B total parameters. LLaDA-UI is trained in two steps. First, we collect large-scale general multimodal data and connect a native dynamic-resolution ViT with LLaDA2.0-mini-base through multimodal pre-training. Because the vision encoder and language backbone are independently pre-trained, this step aligns their representation spaces and equips the diffusion language model with strong visual understanding. We then perform GUI-agent supervised fine-tuning using data accumulated through the UI-Venus project~\citep{gu2025ui,uivenus15}. The training environments cover more than 100 Chinese mobile applications and more than 70 English mobile applications, together with desktop and web scenarios.

% This simple yet effective two-step recipe produces the first practically
% effective block-wise diffusion GUI agent.
Experimental results demonstrate that LLaDA-UI substantially outperforms Qwen2.5-VL-7B and surpasses the similarly scaled autoregressive Qwen3-VL-8B on four of six reported GUI benchmarks. In a controlled paired API evaluation, LLaDA-UI also exhibits consistently lower non-cached inference latency than Qwen3-VL-8B across web, desktop, and mobile inputs. This result suggests a practical inference advantage when neither system benefits from an exact-replay cache hit. The detailed protocol and cache-hit behavior are reported in Section~\ref{sec:paired_latency}. We further analyze the inference behavior of diffusion GUI agents and report not only aggregate benchmark performance but also the settings and scenarios in which LLaDA-UI works or fails.

Our contributions are summarized as follows:
\begin{itemize}[leftmargin=16pt]
    \item Through large-scale multimodal pre-training, we develop a strong LLaDA-based vision--language model that achieves competitive results across representative general multimodal benchmarks.
    \item Through supervised fine-tuning on large-scale GUI data, we develop and open-source the first practical diffusion-based vision-language GUI agent, substantially outperforming Qwen2.5-VL-7B and surpassing Qwen3-VL-8B on four of six grounding and navigation benchmarks.
    \item We provide a detailed empirical analysis of LLaDA-UI, including diffusion-specific inference configurations, paired non-cached API latency, and representative successful and failed scenarios. As the first work to make a diffusion language model operate effectively as a vision-language GUI agent, LLaDA-UI establishes a new possibility for future vision-language agents.
\end{itemize}

\begin{figure*}[tb]
    \centering
    \includegraphics[width=\textwidth]{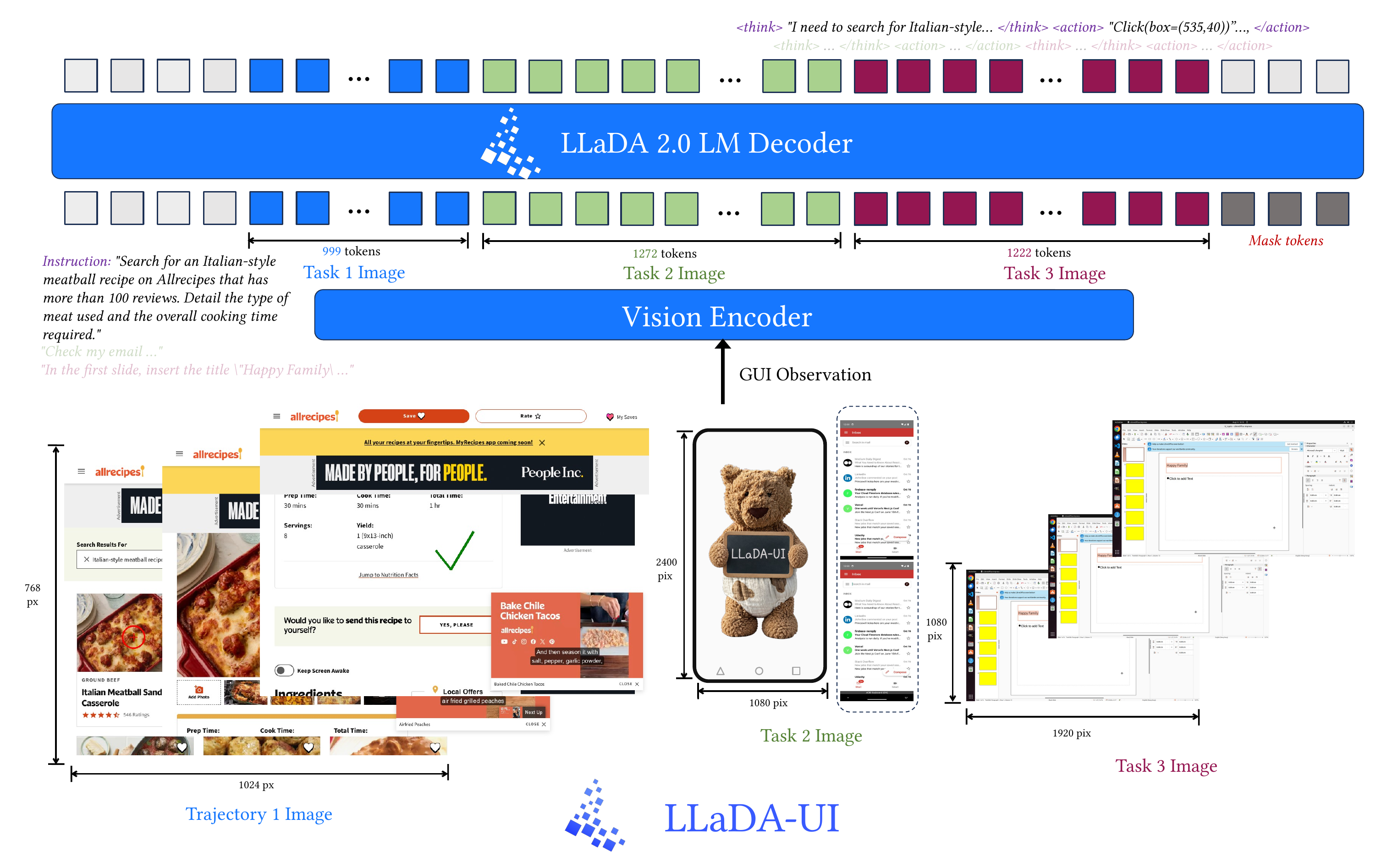}
    \caption{The framework of LLaDA-UI. GUI observations from web, mobile,
    and desktop environments are encoded at native resolution and interleaved
    with task and interaction-history tokens. The LLaDA2.0 block-wise diffusion
    decoder progressively denoises masked output tokens into structured reasoning
    and executable actions.}
    \label{fig:lladavl}
\end{figure*}

\section{LLaDA-UI Architecture}

As shown in Figure \ref{fig:lladavl}, the overall model architecture of LLaDA-UI consists of three components:

\noindent
\textbf{Diffusion Large Language Model:} 
The language model of LLaDA-UI utilizes LLaDA2.0-mini-base~\citep{LLaDA2}, a 16B-parameter MoE language model following an architecture similar to Ling2.0~\citep{Ling2}. Block-wise masked diffusion enables multiple tokens within the current block to be refined in parallel.

\noindent
\textbf{Vision Encoder:}
The vision encoder extracts semantically meaningful representations from both natural images and GUI screenshots. We use pretrained SigLIP~\citep{bai2025qwen2} to initialize the Vision Transformer (ViT). It supports native-resolution inputs and adopts 2D Rotary Positional Embeddings (RoPE) to effectively capture spatial relationships in two-dimensional space.

\noindent
\textbf{Vision-Language Projector:} 
The projector maps visual embeddings into the text embedding space of the large language model, effectively bridging the modality gap between the vision encoder and the language model. Following the design of Qwen2.5-VL~\citep{bai2025qwen2}, to alleviate the computational and inference inefficiency caused by long sequences of image features, we first group and concatenate every four adjacent image features along the spatial dimension, and then project them into the language model’s embedding space using a two-layer multilayer perceptron (MLP). This design substantially reduces computational overhead while providing a flexible mechanism for dynamically compressing image feature sequences of varying lengths.

\section{Pre-training}
\label{sec:pretraining}
LLaDA-UI pre-training connects two independently pre-trained components: a native dynamic-resolution ViT and LLaDA2.0-mini-base as the language backbone.  General multimodal data provides the cross-modal supervision needed to align their representation spaces, enabling visual information to be interpreted and generated through a block-wise diffusion language model. The resulting foundation supplies the perception, OCR, grounding, and reasoning abilities required by the subsequent GUI-Agent SFT stage.
\subsection{Training Recipe}
Within the foundation-building phase of the overall pipeline, multimodal pre-training follows a three-phase curriculum that progresses from cross-modal alignment to perception enhancement and multi-task knowledge injection. The overall setup and objectives of each stage are summarized in Table~\ref{tab:foundational-stages}.

We implement multimodal pre-training with dFactory~\citep{dFactory}, built on
VeOmni~\citep{ma2025veomni}.  The training stack supports distributed
multimodal data loading, native-resolution image processing, sequence packing,
and MoE training.  In particular, sequence packing concatenates short samples
into longer training sequences while preserving sample boundaries in the
objective, improving data throughput and hardware utilization throughout the
three-stage curriculum.

\begin{table}[htbp]
\centering
\small 
\caption{Overview of the foundational pretraining stages, including data composition, token scale, sequence length, and trainable components.}
\label{tab:foundational-stages}
\renewcommand{\arraystretch}{1.0}
\resizebox{\linewidth}{!}{
\begin{tabular}{cccc}
\toprule
\textbf{Stages\&Objective} & \textbf{S0: Vision-Language Alignment} & \textbf{S1: Perception Enhancement}& \textbf{S2: Multi-task Pre-training}   \\
\midrule
\textbf{Data} & Image Caption & \makecell[c]{+ \\ OCR, Grounding \\ Counting, Pure text \\ Interleaved data} & \makecell[c]{+ \\ Multimodal Reasoning \\ Multimodal VQA}  \\
\midrule
\textbf{Tokens} & 5B & 70B & 70B  \\
\midrule
\textbf{Sequence length} & 16K & 16K & 16K\\
\midrule
\textbf{Trainable Params} & Projector & All & All \\
\bottomrule
\end{tabular}
}
\vspace{10pt} 
\end{table}

% 阶段0：视觉-语言对齐。初始阶段（S0）重点在于以提升其与语言模型的对齐程度。该阶段的主要数据来源高质量图像-字幕对、视觉知识集合数据。这些数据集经过精心挑选，旨在培养 ViT 提取有意义视觉表征的能力，使其能够与文本信息有效融合。这种“先对齐”的方法在进入全参数训练之前，为跨模态理解奠定了坚实基础。阶段1：感知增强。在完成初始对齐后，阶段1（S1）转入全参数的多模态预训练。在该阶段，我们解冻模型的所有组件——视觉编码器、融合器以及 LLM——进行联合端到端训练。本阶段引入更细粒度的数据集来增强模型的感知和知识能力，例如图文交错（interleaved）数据、OCR、视觉计数/定位（visual counting/grounding）任务，同时还引入了纯文本数据以保持LLM强大的语言能力。阶段2：多任务预训练。使用多样化的多模态图像数据对模型进行训练，以增强其处理复杂视觉信息的能力。本阶段引入更复杂、对推理要求更高的数据集，例如图文交错（interleaved）数据、多任务学习数据集、视觉问答（VQA）、多模态数学、基于智能体（agent-based）的任务、视频理解以及纯文本数据。这些数据集强化了模型在视觉与语言模态之间建立更深层联系的能力，使其能够应对日益复杂的任务。
\noindent
\textbf{Stage 0: Vision–Language Alignment.} The initial stage (S0) focuses on improving the alignment between the model and the language model. The primary data sources in this stage include high-quality image–caption pairs and visual knowledge collections. These datasets are carefully selected to foster the ViT's ability to extract meaningful visual representations that can be effectively integrated with textual information. This ``alignment-first'' approach establishes a solid foundation for cross-modal understanding before moving on to full-parameter training.

\noindent
\textbf{Stage 1: Perception Enhancement.} After the alignment, Stage 1 (S1) transitions to full-parameter multimodal pre-training. In this phase, we unfreeze all model components, including the vision encoder, the merger, and the LLM for joint end-to-end training. This stage introduces more fine-grained datasets to enhance the model’s perception and knowledge capabilities, such as interleaved image–text data, OCR, and visual counting/grounding tasks. Text-only data is also included to maintain the LLM's strong language abilities.

\noindent
\textbf{Stage 2: Multi-Task Pre-Training.} The model is trained on diverse multimodal image data to improve its capacity to process complex visual information. This stage incorporates more challenging and reasoning-intensive datasets, including interleaved data, multi-task learning datasets, visual question answering (VQA), multimodal mathematics, and agent-based tasks. These datasets strengthen the model’s ability to build deeper connections between visual and linguistic modalities, enabling it to handle increasingly sophisticated tasks.

\subsection{Training Data}

We assemble a broad multimodal mixture containing multilingual image captions,
interleaved image--text documents, OCR, grounding and counting, and multimodal
VQA and reasoning. Captions and OCR annotations are refined with specialist
models, while spatial data from Objects365~\citep{Objects365},
OpenImages~\citep{openimages}, and RefCOCO~\citep{kazemzadeh2014referitgame,yu2016modeling}
are filtered and verified before training. High-quality text from
Ling2.0~\citep{Ling2} and LLaDA2.0~\citep{LLaDA2} is mixed in to preserve general
language, code, and mathematical capabilities. This combination progressively
aligns the two independently pre-trained components and then strengthens
perception and multimodal reasoning.

\subsection{Model Optimization}
\noindent
\textbf{Block Diffusion Language Model Loss.}
The optimization objective of Block Diffusion Language Model~\citep{blockdiffusion} is designed to enable the model to accurately reconstruct the original, uncorrupted tokens within these designated masked blocks using a standard cross-entropy loss. Specifically, we define the training loss under the BDLM paradigm as:
\begin{equation}\label{eq:pretrain_bdlm}
    \mathcal{L}_{\text{BDLM}}(\theta) = - \mathbb{E}_{t, \bm{x}_0, \bm{x_t}} \left[ \frac{\alpha_{t}'}{1-\alpha_{t}}\sum_{k=1}^{K} \sum_{i=1}^{L_B} \mathbb{1}[x_{t,k}^i=\text{[MASK]}] \log p_{\theta}(\bm{x}_{0,k}^i | \bm{x}_{0,<k}, \bm{x}_{t,k}) \right],
\end{equation}
where the expectation is over timestep $t$, the clean sequence $\bm{x}_0$, and its corrupted version $\bm{x}_t$ (tokens masked with probability $1 - \alpha_t$). Indicator $\mathbb{1}[\cdot]$ ensures predictions are made only for masked tokens, and $-\alpha'_t / (1 - \alpha_t)$ is the diffusion-derived time weight. Here $K = L_{\text{total}} / L_B$ is the number of blocks, $L_B$ is the block size, $x^i_{t,k}$ is the $i$-th token in block $k$, $\bm{x}_{0,<k}$ is the preceding clean blocks, and $\bm{x}_{t,k}$ is the noisy version of the current block.

\noindent
\textbf{Load Balancing Strategy.}
In MoE models, imbalanced expert utilization can lead to routing collapse, adversely affecting both computational efficiency and training stability. To address this issue without sacrificing model performance, we adopt an auxiliary-loss-free load balancing mechanism~\citep{deepseekv3}. This method promotes differentiated expert specialization while encouraging a more uniform distribution of computational workload across experts. To further improve numerical stability during training, we apply a scaling operation to the routing gate outputs. Specifically, the gate activations are multiplied by a factor of 2.5, which stabilizes their root-mean-square (RMS) magnitude and prevents excessive variance. For bias updating, we incorporate moderate refinements inspired by prior work~\citep{sjl_moe}. The auxiliary-loss-free bias is updated according to:
\begin{equation}
b_{i}=b_{i}+u\times \frac{(F_{i}-Q_{i})}{\sqrt{\frac{1}{n}\sum_{j=1}^{n}{(F_{j}-Q_{j})^2}}} \,,
\end{equation}
where $F=\mathbb{E}(f)$ denotes the current expert load distribution induced by the bias $b$, and $Q=[\frac{1}{n},\frac{1}{n},\dots,\frac{1}{n}]$ represents the ideal uniform distribution over $n$ experts. By applying RMSNorm-style normalization to the expert load imbalance term, the bias updates are smoothed, leading to more stable and effective load balancing throughout training.

\noindent
\textbf{Mask-Token Reweighting.}
Multimodal samples vary substantially in target length.  Token averaging can
allow long samples to dominate the gradient, whereas sample averaging can
overemphasize short responses.  We balance these regimes with an inverse
square-root weight based on the number of active masked targets:
\begin{equation}\label{eq:mtrs}
    \mathcal{L}_{\text{MTRS}}
    = \frac{\sum_j \beta_j\mathcal{L}_{\text{BDLM}}^{(j)}}{\sum_j\beta_j},
    \qquad
    \beta_j = \frac{1}{\sqrt{\sum_{k=1}^{K}\sum_{i=1}^{L_B}
    \mathbb{1}[x_{t,k}^{i,(j)}=\text{[MASK]}]}}.
\end{equation}
This objective moderates length-related gradient imbalance while retaining the
ability to learn from both compact and long-form multimodal targets.

\noindent
\textbf{Complementary Masking.}
For each clean target, complementary masking constructs two corrupted views
whose masks are logical inverses.  Every token is therefore observed once and
predicted once across the pair, improving information utilization and reducing
token-level sampling bias during multimodal pre-training.

\section{GUI-Agent Supervised Fine-Tuning}
We specialize the model with supervised GUI trajectories that map a task, current screenshot, and interaction history to a structured response containing reasoning and executable actions. 

\subsection{Training Data}
The corpus covers four complementary domains. Mobile navigation spans more
than 100 Chinese apps and more than 70 English apps; desktop and web data cover
cross-application and browser workflows; and grounding data connects language
instructions to visual targets. After conversion, filtering, and deduplication,
the final GUI-agent SFT mixture contains more than 6M samples in total.

\subsection{Data Collection and Processing}
\label{sec:data_collection}
\begin{figure}[tbp]
    \centering
    \includegraphics[width=\linewidth]{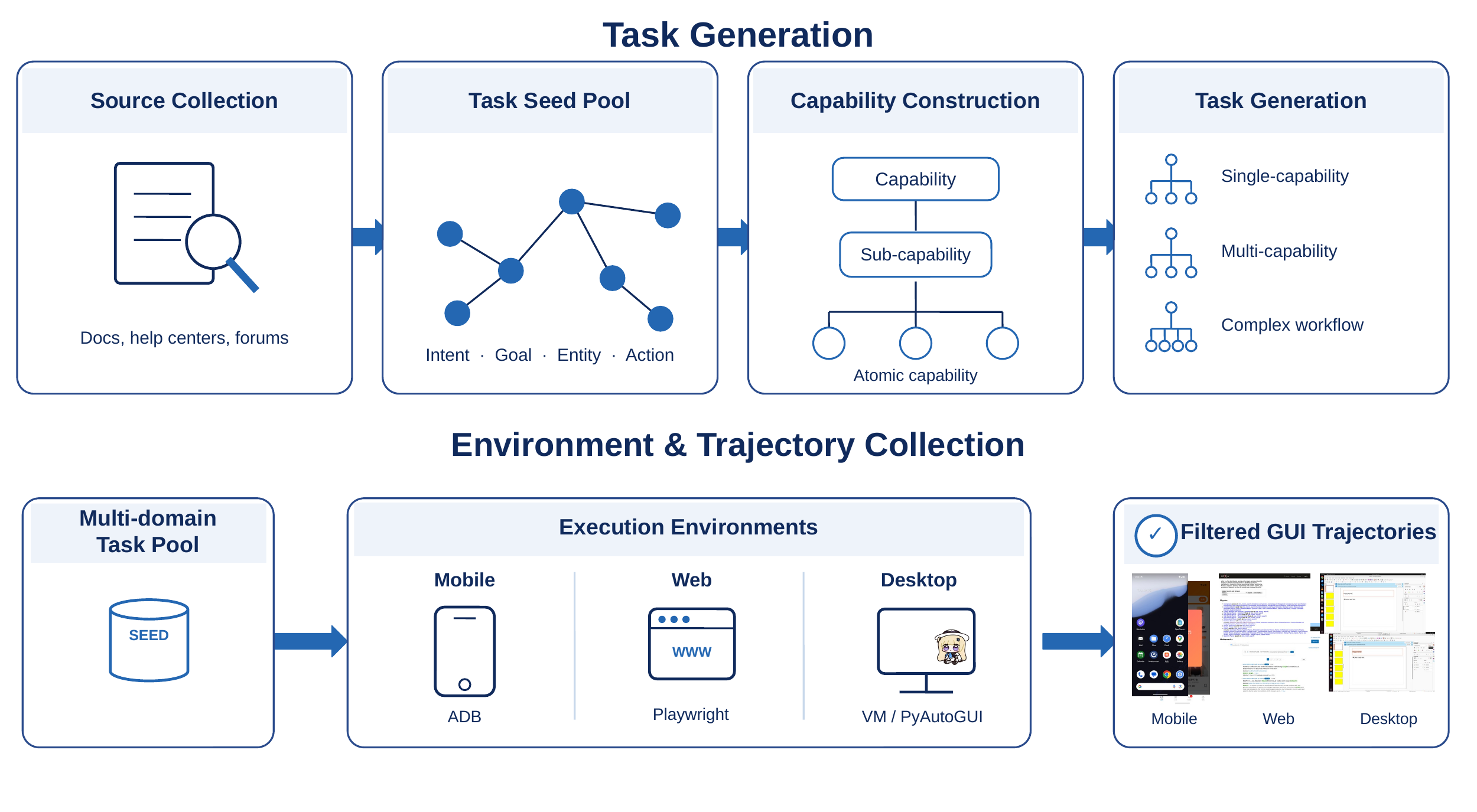}
    \caption{Overview of the GUI data-generation pipeline. Task posing is
    followed by sub-capability decomposition, capability composition,
    environment execution, and validation into training trajectories.}
    \label{fig:gui_data_pipeline}
\end{figure}

Figure~\ref{fig:gui_data_pipeline} summarizes the collection process. We first pose diverse tasks for each environment and decompose them into a hierarchy of atomic GUI capabilities. Compatible capabilities are then composed into increasingly complex tasks, which are executed in instrumented mobile, desktop, and web environments. The resulting trajectories are validated and converted to the common training format, and grounding examples are prepared through the same filtering and normalization pipeline.

Each navigation target uses a tagged
\texttt{<think>...</think><action>...</action>} response, whereas grounding
returns the target point directly. The full prompt templates for grounding,
mobile, desktop, and web are provided in Appendix~\ref{app:prompts}.

\subsection{Training Recipe}

GUI-agent SFT retains the block diffusion objective and MoE balancing mechanism used during pre-training. We initialize from the multimodal foundation and fine-tune the full model for three epochs with a peak learning rate of $1\times10^{-5}$. We use cosine decay with a 3\% warmup ratio, a maximum sequence length of 16K tokens, mixed-precision training, and a micro-batch size of one per GPU. {Training runs use 32--64 H100 GPUs depending on the data mixture, with a maximum image-pixel budget of approximately 12.8M pixels.}

\section{Experiments}
\definecolor{tbarc}{HTML}{F5F9FF}
\definecolor{tbdiff}{HTML}{FFF8F3}

We evaluate the multimodal foundation before GUI fine-tuning and then evaluate
LLaDA-UI on grounding and dynamic navigation. This separation identifies
the capabilities established by multimodal pre-training and those introduced
by GUI-agent SFT.

\begin{table*}[!tb]
\centering
\scriptsize
\renewcommand{\arraystretch}{0.94}
\setlength\tabcolsep{1.2pt}
\caption{General multimodal and text results before GUI-agent SFT. AR and
diffusion models are shown in \colorbox{tbarc}{\scriptsize blue} and
\colorbox{tbdiff}{\scriptsize orange}, respectively.}
\label{tab:pretrain_results}
\resizebox{\textwidth}{!}{%
\begin{tabular}{l >{\columncolor{tbarc}}c >{\columncolor{tbarc}}c >{\columncolor{tbarc}}c >{\columncolor{tbarc}}c >{\columncolor{tbdiff}}c >{\columncolor{tbdiff}}c >{\columncolor{tbdiff}}c}
\toprule
& \multicolumn{4}{c}{\cellcolor{tbarc}\textbf{Autoregressive Models}}
& \multicolumn{3}{c}{\cellcolor{tbdiff}\textbf{Diffusion Models}} \\
\cmidrule(lr){2-5}\cmidrule(lr){6-8}
& \textbf{Qwen2.5-VL}
& \textbf{Qwen2.5-VL}
& \textbf{Qwen3-VL}
& \textbf{Qwen3-VL}
& \textbf{LLaDA-V}
& \textbf{SDAR-VL}
& \textbf{LLaDA-UI} \\
\midrule
Architecture & Dense & Dense & Dense & Dense & Dense & Dense & MoE \\
Base LLM & Qwen2.5 & Qwen2.5 & Qwen3 & Qwen3 & LLaDA & Qwen3 & LLaDA2.0 \\
% LLM params & 3B & 7B & 2B & 8B & 8B & 8B & 16.7B total \\
Pre-train tokens & 4.1T & 4.1T & 2.2T & 2.2T & -- & -- & 145B \\
\midrule
\multicolumn{8}{l}{\textbf{Reasoning}} \\
$\mathrm{MMMU}_{\mathrm{val}}$ & 46.4 & 51.3 & 53.4 & \bestscore{69.6} & 48.6 & 53.0 & \underline{56.2} \\
$\mathrm{MMMUPro}_{\mathrm{standard}}$ & 31.1 & 36.3 & \underline{36.5} & \bestscore{55.9} & 35.2 & 35.1 & 36.0 \\
$\mathrm{MathVista}_{\mathrm{mini}}$ & 60.2 & 68.6 & 61.3 & \bestscore{77.2} & 59.7 & 62.5 & \underline{70.5} \\
WeMath & 18.4 & 33.3 & -- & -- & 24.7 & \underline{33.9} & \bestscore{34.6} \\
$\mathrm{MathVision}_{\mathrm{mini}}$ & 21.0 & 25.2 & \underline{31.6} & \bestscore{53.9} & 21.2 & 24.2 & 27.5 \\
$\mathrm{MathVerse}_{\mathrm{mini}}$ & 47.6 & 49.2 & \underline{52.1} & \bestscore{62.1} & 29.1 & 36.6 & 48.0 \\
$\mathrm{MathVerse}_{\mathrm{vision\_only}}$ & 35.6 & \bestscore{42.7} & -- & -- & 22.5 & 36.5 & \underline{39.7} \\
% LogicVista & 40.3 & 44.1 & 35.8 & \textbf{55.3} & 35.1 & 38.7 & \textbf{40.7} \\
\midrule
\multicolumn{8}{l}{\textbf{General}} \\
SimpleVQA & 38.6 & \underline{44.3} & 40.7 & \bestscore{50.2} & 25.5 & 33.8 & \underline{44.3} \\
HallusionBench & 46.6 & \underline{51.9} & 51.4 & \bestscore{61.1} & 50.2 & 44.4 & 50.7 \\
$\mathrm{MMBench}_{\mathrm{en}}$ & 79.1 & \underline{83.5} & \underline{83.5} & \bestscore{84.5} & 82.9 & 82.2 & 80.5 \\
$\mathrm{MMBench}_{\mathrm{cn}}$ & 78.1 & \underline{83.4} & 75.9 & \bestscore{84.7} & 79.2 & 78.4 & 77.7 \\
MMStar & 55.9 & \underline{63.9} & 58.3 & \bestscore{70.9} & 60.1 & 59.9 & 61.3 \\
RealWorldQA & 65.4 & \underline{68.5} & 63.9 & \bestscore{71.5} & 63.2 & 66.5 & 63.5 \\
\midrule
\multicolumn{8}{l}{\textbf{OCR and Chart}} \\
ChartVQA & 83.4 & 84.1 & 79.1 & \bestscore{89.6} & 82.7 & 82.7 & \underline{84.8} \\
DocVQA & 92.7 & \underline{94.9} & 93.3 & \bestscore{96.1} & 83.9 & 88.3 & 91.5 \\
InfoVQA & 75.6 & \underline{81.7} & 72.4 & \bestscore{83.1} & 66.3 & 73.2 & 71.5 \\
CharXiv (DQ) & 58.6 & \underline{73.9} & 62.3 & \bestscore{83.0} & 47.0 & 66.5 & 72.2 \\
CharXiv (RQ) & 31.3 & \underline{42.5} & 31.6 & \bestscore{46.4} & 28.9 & 33.3 & 35.6 \\
OCRBench & 792 & 842 & \underline{858} & \bestscore{896} & 632 & 726 & 855 \\
$\mathrm{AI2D}_{\mathrm{w\ mask}}$ & 78.6 & \underline{82.6} & 76.9 & \bestscore{85.7} & 77.8 & 79.9 & 80.2 \\
\midrule
% \multicolumn{8}{l}{\textbf{Multi-image}} \\
% BLINK & 47.6 & 56.4 & 53.8 & \textbf{69.1} & 52.4 & \textbf{52.9} & 52.5 \\
% MUIRBench & 47.7 & 59.6 & 47.4 & \textbf{64.4} & 48.3 & \textbf{50.2} & 49.1 \\
% \midrule
\multicolumn{8}{l}{\textbf{Other}} \\
CountBench & 72.5 & \bestscore{86.4} & 84.1 & 80.5 & 74.3 & 75.4 & \underline{86.2} \\
VLRewardBench & 42.1 & \underline{49.7} & -- & -- & 45.9 & 39.0 & \bestscore{50.8} \\
% \midrule
% \multicolumn{8}{l}{\textbf{Text-centric}} \\
% MMLU & 66.5 & 73.1 & 69.7 & \textbf{84.6} & 57.5 & \textbf{77.6} & 70.2 \\
% MMLU-Pro & 40.0 & 46.0 & 46.0 & \textbf{72.0} & 28.7 & \textbf{55.9} & 46.1 \\
% DROP & 64.0 & 73.6 & 68.4 & \textbf{87.9} & 59.3 & 77.3 & \textbf{79.1} \\
% HumanEval & 61.6 & 66.5 & 66.5 & \textbf{92.7} & 17.7 & 54.3 & \textbf{76.2} \\
% MBPP & 56.2 & 65.1 & 65.3 & \textbf{87.4} & 42.8 & \textbf{75.6} & 73.8 \\
% GSM8K & 80.7 & 86.1 & 85.9 & \textbf{95.6} & 73.0 & 88.6 & \textbf{88.9} \\
% MATH & 61.8 & 63.7 & 78.7 & \textbf{94.7} & 31.9 & 68.1 & \textbf{68.3} \\
\bottomrule
\end{tabular}}
\end{table*}

\subsection{Pre-training Evaluation}

\paragraph{Data sources and benchmarks.}
The evaluation covers \textbf{multimodal reasoning} (MMMU \cite{yue2024mmmu}, MMMU-Pro \cite{yue2025mmmupro}, MathVista \cite{mathvista}, We-Math \cite{We-Math}, MathVision \cite{wang2024measuring}, and MathVerse \cite{zhang2024mathverse}), \textbf{general visual question answering} (SimpleVQA \cite{SimpleVQA}, HallusionBench \cite{HallusionBench}, MMBench \cite{MMBench}, MMStar \cite{MMStar}, and RealWorldQA \cite{RealWorldQA}), \textbf{OCR and document understanding} (ChartVQA \cite{Chartqa}, DocVQA \cite{DoCVQA}, InfoVQA \cite{InfoVQA}, CharXiv \cite{CharXiv}, OCRBench \cite{OCRBench}, and AI2D \cite{AI2D}), \textbf{counting} (CountBench \cite{CountBench}), and \textbf{preference evaluation} (VLRewardBench \cite{VLRewardBench}).

% , and \textbf{text-centric reasoning} (MMLU \cite{MMLU}, MMLU-Pro \cite{MMLU-Pro}, DROP \cite{DROP}, HumanEval \cite{HumanEval}, MBPP \cite{MBPP}, GSM8K \cite{GSM8K}, MATH \cite{MATH}).

% We use the standard public splits and metrics of each benchmark; split-specific results such as $\mathrm{MMMU}_{\mathrm{val}}$ and $\mathrm{MMMUPro}_{\mathrm{standard}}$ are reported explicitly in Table~\ref{tab:pretrain_results}.

\paragraph{Implementation details.}
The evaluated model is the multimodal foundation produced by the curriculum in
Section~\ref{sec:pretraining}, before GUI-agent SFT. Evaluation uses the same
image preprocessing and block-wise diffusion decoder across the benchmark
suite.

\paragraph{Baselines.}
We compare with Qwen2.5-VL~\citep{bai2025qwen2} and
Qwen3-VL~\citep{team2025qwen3VL}, as well as the diffusion VLMs
LLaDA-V~\citep{you2025llada} and SDAR-VL~\citep{cheng2025sdar}.

\paragraph{Results.}
Table~\ref{tab:pretrain_results} reports the complete image and text evaluation from the LLaDA-UI study. The results show that the foundation checkpoint (before GUI-agent SFT) is trained on only 145B tokens, which is competitive with Qwen2.5-VL's 4.1T and Qwen3-VL's 2.2T, yet establishes a new state of the art among diffusion-based MLLM baselines. Notably, our advantages are more pronounced in reasoning and text-rich tasks: compared to SDAR-VL-8B, our model achieves a performance lead of 12.8\%/13.6\%/31.1\% on MathVista, MathVision, and MathVerse, respectively, while surpassing it by 2.5\%/8.6\%/17.8\% on ChartQA, CharXiv-DQ, and OCRBench.

\subsection{GUI-Agent Evaluation}

\paragraph{Data sources and benchmarks.}
{ScreenSpot-V2 \cite{wu2025atlas} and ScreenSpot-Pro \cite{li2025screenspot} evaluate static GUI grounding; AndroidWorld \cite{rawles2025androidworld} and MobileWorld \cite{kong2026mobileworld} evaluate dynamic mobile navigation; OSWorld-Verified \cite{xie2026scaling} evaluates desktop navigation; and WebVoyager \cite{he-etal-2024-webvoyager} evaluates web navigation. Grounding uses point-in-box accuracy, while navigation benchmarks report end-to-end task success in executable environments.}

\paragraph{Implementation details.}
All agents receive the same task: screenshot observation, interaction history, action space, environment feedback, and action budget whenever supported. Spatial outputs are normalized to $[0,1000]$, parsed into platform actions, and executed by the corresponding benchmark runner. We validate both the standalone Hugging Face inference path and the OpenAI-compatible SGLang serving path end-to-end on a two-GPU host, and the latter uses two independent data-parallel workers.

\paragraph{Baselines.}
We compare with GPT-4o \citep{gpt4o}, UI-TARS \citep{qin2025ui}, GUI-G$^2$ \citep{tang2026gui}, GUI-Owl \citep{ye2025mobile}, UI-TARS-1.5 \citep{Ui-tars-1.5}, UI-Venus-1.0 \citep{gu2025ui}, UI-Venus-1.5 \citep{team2026ui}, Holo2 \citep{hai2025holo2modelfamily}, Step-GUI \citep{yan2025step}, MAI-UI \citep{zhou2025mai}, Qwen2.5-VL-7B~\citep{bai2025qwen2},
Qwen3-VL-8B~\citep{team2025qwen3VL}, and Qwen3.5-9B \cite{qwen3.5} under the corresponding
evaluation protocols.

\subsubsection{Overall Results}

\begin{table}[htbp]
\centering
\small
\caption{GUI-agent results (\%). ScreenSpot-V2 and ScreenSpot-Pro evaluate
grounding; AndroidWorld and MobileWorld evaluate mobile navigation;
OSWorld-Verified evaluates desktop navigation; and WebVoyager evaluates web
navigation. LLaDA-UI's WebVoyager score uses \texttt{gpt-4o-2024-11-20} as
the judge. Bold and underlined values indicate the best and second-best
reported results in each column, respectively.}
\label{tab:agent_main}
\resizebox{\linewidth}{!}{%
\begin{tabular}{lcccccc}
\toprule
& \multicolumn{2}{c}{\textbf{Grounding}} & \multicolumn{2}{c}{\textbf{Mobile}} & \textbf{Desktop} & \textbf{Web} \\
\cmidrule(lr){2-3}\cmidrule(lr){4-5}\cmidrule(lr){6-6}\cmidrule(lr){7-7}
\textbf{Model} & \textbf{SS-V2} & \textbf{SS-Pro} & \textbf{AndroidWorld} &
\textbf{MobileWorld} & \textbf{OSWorld-V} & \textbf{WebVoyager} \\
\midrule
\rowcolor{myBlue!12}
\multicolumn{7}{l}{\textbf{General-purpose vision-language models}} \\
GPT-4o & 20.1 & 0.8 & 30.6 & - & - & -\\
Qwen2.5-VL-7B & 86.0 & 26.8 & 25.5 & 7.0 & 3.0 & 11.0 \\
Qwen3-VL-8B & 93.0 & 52.7 & 47.9 & 9.4 & \underline{33.9} & 45.2 \\
Qwen3.5-9B & \underline{94.0} & \guibestscore{65.2} & 57.8 & \underline{17.9} & \guibestscore{41.8} & 46.6 \\
\midrule
\rowcolor{myBlue!12}
\multicolumn{7}{l}{\textbf{Specialized GUI agents}} \\
% Claude Computer Use & - & 17.1 & -  & - & - & -\\
% SeeClick-9.6B & 55.1 & 1.1 & -  & - & - & -\\
% FOCUS-2B & - & 13.3 & -  & - & - & -\\
% CogAgent-18B & - & 7.7 & -  & - & - & -\\
% Aria-UI & - & 11.3 & -  & - & - & -\\
% OS-Atlas-7B & 84.1 & 18.9 & -  & - & - & -\\
% ShowUI-2B & 77.3 & 7.7 & -  & - & - & -\\
% UGround-7B & 76.3 & 16.5 & -  & - & - & -\\
% UGround-V1-7B & - & 31.1 & -  & - & - & -\\
UI-TARS-7B & 91.6 & 35.7 & -  & - & - & -\\
% JEDI-7B & 91.7 & 39.5 & -  & - & - & -\\
% GUI-Actor-7B & 92.1 & 44.6 & -  & - & - & -\\
% UI-R1-E-3B & 89.5 & 33.5 & -  & - & - & -\\
% InfiGUI-R1-3B & - & 35.7 & -  & - & - & -\\
% GUI-G1-3B & - & 37.1 & -  & - & - & -\\
% SE-GUI-7B & 90.3 & 47.3 & -  & - & - & -\\
% Phi-Ground-7B-16C-DPO & 83.8 & 43.2 & -  & - & - & -\\
GUI-G$^2$-7B & 93.8 & 47.5 & -  & - & - & -\\
% Aguvis-7B & 80.5 & - & -  & - & - & -\\
% Aguvis-7B & - & - & 44.0 & - & - & -\\
% SeedVL-1.5 & - & - & 62.1 & - & - & -\\
% GUI-Critic-R1-7B & - & - & 27.6 & - & - & -\\
% UGround & - & - & 44.0 & - & - & -\\
% Aria-UI & - & - & 44.8 & - & - & -\\
% GLM-4.5v & - & - & 57.0 & - & - & -\\
% OpenCUA-7B & 92.3 & 50.0 & - & - & - & -\\
% GTA1-7B & 92.4 & 50.1 & - & - & - & -\\
GUI-Owl-7B & 92.8 & 54.9 & 66.4 & 7.7 & - & -\\
UI-TARS-1.5-7B & 91.6 & 35.7 & 30.0 & - & - & -\\
UI-Venus-1.0-7B & \guibestscore{94.1} & 50.8 & 49.1 & 8.5 & - & - \\
UI-Venus-1.5-2B & 92.8 & 57.7 & 55.6 & - & - & 56.4\\
Holo2-8B & 93.2 & 58.9 & 60.4 & - & - & \guibestscore{80.2}\\
Step-GUI-4B & 93.6 & \underline{60.0} & \underline{63.9} & - & - & 47.8\\
MAI-UI-2B & 92.5 & 57.4 & 49.1 & - & - & -\\
\midrule
\rowcolor{myBlue!12}
\multicolumn{7}{l}{\textbf{Diffusion GUI agent}} \\
LLaDA-UI & 90.7 & 52.9 & 53.5 &
\guibestscore{25.6} & 29.4 & \underline{56.9} \\
\bottomrule
\end{tabular}}
\end{table}

LLaDA-UI exceeds Qwen2.5-VL-7B on every reported benchmark. Relative to
Qwen3-VL-8B, it is stronger on ScreenSpot-Pro, AndroidWorld, MobileWorld, and
WebVoyager, while ScreenSpot-V2 is close and OSWorld-Verified remains the main
gap. These results show that block-wise diffusion generation can support both
structured grounding and long-horizon interaction across multiple platforms.

\subsubsection{Qualitative Diffusion Decoding}
{Figure~\ref{fig:combined_diffusion_cases} shows two complete AndroidWorld
denoising sequences.} Each case retains the uncropped observation and full
model output, including reasoning, action tags, and special tokens.

\begin{figure}[H]
\centering
\animategraphics[width=0.62\linewidth,autoplay,loop,poster=last]{3}%
  {figs/diffusion_cases/frames/combined_cases/frame-}{1}{132}
\caption{Two complete LLaDA-UI diffusion-decoding cases. The first sequence
forms a numeric entry action, and the second grounds a list item before
producing the click. Best viewed in Adobe Acrobat Reader.}
\label{fig:combined_diffusion_cases}
\end{figure}

\subsubsection{Ablation Studies}
\label{sec:inference_ablation}

We study EOS handling and the joint choice of block size and denoising steps on
an intermediate LLaDA-UI checkpoint, holding the task split, prompt, action
parser, and maximum environment steps fixed.

\begin{table}[H]
\centering
\small
\caption{AndroidWorld decoding ablation on an intermediate LLaDA-UI
checkpoint.}
\label{tab:agent_inference_ablation}
\resizebox{\linewidth}{!}{%
\begin{tabular}{lccccc}
\toprule
\textbf{Decode setting} & \textbf{EOS policy} & \textbf{Block} &
\textbf{Steps} & \textbf{Success} & \textbf{Avg. actions} \\
\midrule
Baseline & Early stop & 32 & 32 & 42.7\% (50/116) & 14.9 \\
Two-stage decoding & Early stop & 32 & 32 & 50.9\% (58/115) & 13.7 \\
No-EOS & Disabled & 32 & 32 & \bestscore{52.6\% (61/116)} & 15.8 \\
Larger block/fewer steps & Default & 64 & 16 & 33.0\% (38/115) & 15.1 \\
\bottomrule
\end{tabular}}
\end{table}

Disabling EOS early stopping improves success by 9.9 percentage points in this
controlled comparison. This suggests that early
termination can interrupt the formation of a parseable action, but the setting
should still be validated for the released model. Block size and denoising
steps jointly affect decoding reliability: moving from block 32 with 32 steps
to block 64 with 16 steps reduces success sharply from 42.7\% to 33.0\%.
Coarsening blocks while reducing refinement rounds therefore causes a
substantial loss in interactive reliability.

\subsubsection{Paired API Latency}
\label{sec:paired_latency}

We additionally conduct a paired latency study against Qwen3-VL-8B. Both
models are evaluated on four H100 GPUs with a maximum generation length of 2,048 tokens. We measure model API wall-clock time only, excluding environment reset and action execution, and enable normal EOS stopping for both models. The
primary comparison uses five non-cache-hit calls per model and domain. Exact Qwen request replays that hit its serving cache are excluded; visually equivalent screenshots with different PNG hashes are used to complete its non-cached sample sets.

\begin{table}[htbp]
\centering
\small
\caption{Model API latency over five non-cache-hit calls per domain. Each cell
reports mean / median / range in seconds. These fixed-input measurements
characterize per-call latency rather than throughput.}
\label{tab:paired_latency}
\resizebox{\linewidth}{!}{%
\begin{tabular}{llrrcrr}
\toprule
\textbf{Domain} & \textbf{Model} & \textbf{Mean} & \textbf{Median} &
\textbf{Range} & \textbf{Tokens} & \textbf{Mean speedup} \\
\midrule
Web & LLaDA-UI & 4.764 & 4.801 & 4.474--5.156 & 144 & 3.579$\times$ \\
& Qwen3-VL-8B & 17.050 & 16.458 & 15.873--20.044 & 55.2 & -- \\
\addlinespace
OSWorld & LLaDA-UI & 6.379 & 6.313 & 5.911--6.741 & 129 & 6.826$\times$ \\
& Qwen3-VL-8B & 43.545 & 42.787 & 42.083--47.330 & 70.0 & -- \\
\addlinespace
MobileWorld & LLaDA-UI & 5.921 & 5.944 & 5.434--6.299 & 67 & 8.951$\times$ \\
& Qwen3-VL-8B & 53.000 & 52.394 & 51.882--54.298 & 65.2 & -- \\
\bottomrule
\end{tabular}}
\end{table}

As shown in Table \ref{tab:paired_latency}, LLaDA-UI is 3.579--8.951$\times$ faster by mean latency; the corresponding
median speedups are 3.428$\times$, 6.778$\times$, and 8.814$\times$ on Web,
OSWorld, and MobileWorld. Qwen3-VL-8B also exhibits a separate exact-replay
cache-hit mode (0.890--1.284 s on Web, 1.444 s on OSWorld, and 1.746--1.794 s
on MobileWorld), whereas the evaluated LLaDA-UI SGLang stack exposes no
comparable cache-hit path. We report those values separately because normal
GUI interaction produces a new screenshot at each step. 
%One incomplete Qwen mobile stream is excluded; only its successful retry API call enters thenon-cached sample set.

\subsubsection{Structured Actions and Long-Horizon Failures}
\label{sec:action_validity}

We analyze a complete 116-task AndroidWorld evaluation from an intermediate
LLaDA-UI checkpoint, containing 1,849 model steps. These diagnostics are
separate from the final benchmark results in Table~\ref{tab:agent_main}.
LLaDA-UI follows the requested structured instruction format reliably: 98.65\%
of responses parse, 98.59\% satisfy the action schema, and all 1,457
coordinate-bearing actions lie in the valid range. Most residual format errors
are multiple action calls in a response rather than malformed coordinates.

\begin{table}[htbp]
\centering
\small
\caption{Structured-action validity on AndroidWorld for an intermediate
LLaDA-UI checkpoint. These diagnostic values are not the final benchmark
result.}
\label{tab:action_validity}
\begin{tabular}{lrr}
\toprule
\textbf{Metric} & \textbf{Rate} & \textbf{Count} \\
\midrule
Task success & 51.72\% & 60/116 tasks \\
Parse rate & 98.65\% & 1,824/1,849 steps \\
Schema-valid rate & 98.59\% & 1,823/1,849 steps \\
Coordinate-valid rate & 100.00\% & 1,457/1,457 actions \\
Exact-repetition rate & 18.82\% & 348/1,849 steps \\
Multiple-action rate & 1.30\% & 24/1,849 steps \\
\bottomrule
\end{tabular}
\end{table}

Structural validity alone does not explain task failure. Exact repetition rises
from 5.41\% of steps in successful trajectories to 26.70\% in failed ones, and
success declines from 62.96\% on tasks with at most five optimal actions to
14.29\% beyond 20 actions. These trends indicate that long trajectories often
degrade through repeated actions and an inability to escape an unproductive UI
state. On OSWorld, a different bottleneck dominates: screenshots are much
larger and contain smaller targets, exposing insufficient high-resolution
perception and inaccurate pointing. Thus, the model generally follows the
required instruction and action format, while long-horizon recovery and
high-resolution grounding remain the primary failure modes.

\begin{table}[htbp]
\centering
\small
\caption{AndroidWorld success for the same intermediate LLaDA-UI checkpoint,
stratified by annotated optimal task length.}
\label{tab:success_by_optimal_length}
\begin{tabular}{lrrr}
\toprule
\textbf{Optimal actions} & \textbf{Tasks} & \textbf{Successes} & \textbf{Success rate} \\
\midrule
1--5 & 54 & 34 & 62.96\% \\
6--10 & 35 & 17 & 48.57\% \\
11--20 & 20 & 8 & 40.00\% \\
20+ & 7 & 1 & 14.29\% \\
\bottomrule
\end{tabular}
\end{table}

\subsubsection{Cross-Platform Interaction Traces}
Figures~\ref{fig:web_trace_animation}--\ref{fig:mobile_trace_animation}
present three complete, successful trajectories across WebVoyager, OSWorld,
and MobileWorld.  Every
animation frame pairs the current GUI observation with the model's verbatim
\texttt{<think>} and \texttt{<action>} output for that step.  The sequence
shows long-horizon constraint tracking on the web, formatted content transfer
between desktop applications, and cross-application information use on mobile.

\begin{figure*}[t]
\centering
\animategraphics[width=0.92\linewidth,autoplay,loop,poster=first]{0.6}%
  {figs/agent_traces/animated/webvoyager_google_flights/frame-}{1}{20}
\caption{Complete 20-step WebVoyager trajectory for constrained flight search.
Each frame retains the full recorded reasoning and executable action.  GIFs are
best viewed in Adobe Acrobat Reader.}
\label{fig:web_trace_animation}
\end{figure*}

\begin{figure*}[t]
\centering
\animategraphics[width=0.92\linewidth,autoplay,loop,poster=first]{0.6}%
  {figs/agent_traces/animated/osworld_calc_to_writer/frame-}{1}{10}
\caption{Complete 10-step OSWorld trajectory for transferring a formatted Calc
table to Writer and saving the resulting document.  Each frame retains the full
recorded reasoning and executable action.  GIFs are best viewed in Adobe
Acrobat Reader.}
\label{fig:os_trace_animation}
\end{figure*}

\begin{figure*}[t]
\centering
\animategraphics[width=0.88\linewidth,autoplay,loop,poster=first]{0.6}%
  {figs/agent_traces/animated/mobileworld_email_to_alarm/frame-}{1}{12}
\caption{Complete 12-step MobileWorld trajectory for reading an event time from
email and setting an alarm one hour earlier.  Each frame retains the full
recorded reasoning and executable action.  GIFs are best viewed in Adobe
Acrobat Reader.}
\label{fig:mobile_trace_animation}
\end{figure*}

\section{Conclusion}

We presented LLaDA-UI, a practically effective block-wise diffusion vision--language GUI agent. LLaDA-UI follows a two-stage recipe: general multimodal pre-training aligns a native-resolution ViT with the LLaDA2.0-mini-base diffusion language backbone, and GUI-agent SFT introduces executable behavior across mobile, desktop, web, and grounding tasks. LLaDA-UI substantially outperforms Qwen2.5-VL-7B and surpasses the similarly scaled autoregressive Qwen3-VL-8B on four of six reported grounding and cross-platform navigation benchmarks. As the first open-source diffusion GUI agent, it demonstrates that diffusion language models offer a viable new direction for vision--language agents.

\paragraph{Limitations.}
First, LLaDA-UI can repeat actions and enter unproductive loops on long-horizon tasks. This behavior may partly reflect model capacity and is also frequently observed in GUI agents below the 100B-parameter scale. Second, its grounding and perception degrade on very large screenshots with small targets, as exposed most clearly by OSWorld. Third, interactive performance remains sensitive to the block size and number of denoising steps: small changes to these inference hyperparameters can materially alter both task success and latency. Improving long-horizon recovery, high-resolution perception, and decoding robustness are therefore the main directions for future work.

\appendix
\section{GUI-Agent Prompt Templates}
\label{app:prompts}

The following templates expose the released prompt structure. Values in braces
are replaced by the task and current visual observation at inference time. For
MobileWorld, OSWorld, and WebVoyager, we use a current-image-only multi-turn
protocol. At the first step, the request contains a system message followed by
a user message with the current screenshot. At every later step, the role
sequence is
\texttt{system}, \texttt{user("")}, \texttt{assistant(previous)}, and
\texttt{user(current)}. The assistant message preserves the latest raw
\texttt{<think>...</think><action>...</action>} output. The final user
message places text before the image. Thus, every request contains exactly one
image---the current screenshot---and never replays a historical screenshot.
The desktop template below omits machine-specific credentials. Grounding uses
the single-turn template shown separately.

\lstdefinestyle{venusprompt}{
  basicstyle=\ttfamily\fontsize{6.2}{6.7}\selectfont,
  breaklines=true,
  breakatwhitespace=true,
  columns=fullflexible,
  frame=single,
  rulecolor=\color{secondcolor!35},
  backgroundcolor=\color{secondcolor!4},
  xleftmargin=2pt,
  xrightmargin=2pt,
  aboveskip=5pt,
  belowskip=8pt,
  showstringspaces=false
}

\subsection{Grounding}
\begin{lstlisting}[style=venusprompt]
Output the center point of the position corresponding to the following instruction:
{instruction}

The output should just be the coordinates of a point, in the format [x,y]. Additionally, if the task is infeasible (e.g., the task is not related to the image), the output should be [-1,-1].

{current screenshot}
\end{lstlisting}

\subsection{Mobile (MobileWorld)}
\begin{lstlisting}[style=venusprompt]
You are a GUI Agent. Your role is to analyze the user's task, provide clear and accurate answers, and execute the task with precise actions.

AVAILABLE ACTIONS
Click(box=(x,y)): tap the coordinate.
Drag(start=(x1,y1), end=(x2,y2)): long-press and drag.
Swipe(start=(x1,y1), end=(x2,y2)): swipe or scroll.
DoubleClick(box=(x,y)): double tap the coordinate.
LongPress(box=(x,y)): long-press the coordinate.
Type(content=''): enter text in the active field.
LaunchApp(app=''): launch the target app.
Wait(): wait for the page, animation, or content to load.
CallUser(content=''): request user takeover or information.
GetScreenshot(): save a screenshot to the device album.
PressBack(), PressHome(), PressEnter(), PressRecent(): invoke the corresponding system operation.
Answer(content=''): answer the user's question.
Finished(content=''): mark the task complete and report status.
All screen coordinates range from (0,0) at the top left to (999,999) at the bottom right.

INSTRUCTIONS
- Understand the task goal before acting.
- Carefully examine the current screenshot; a summarized history can over-claim effects.
- Use CallUser when additional information or takeover is required.
- Explore hidden content with Swipe in different directions when necessary.
- To copy text, select the exact text and click copy in the selection bar.
- To paste, long-press the target text box and click paste in the selection bar.

OUTPUT FORMAT
<think>your thinking process</think>
<action>the next action</action>

CURRENT USER MESSAGE (text first, current image last)
### User Task
{task}

### Previous Actions
{empty in current-image-only evaluation}

### Current Screenshot
{current screenshot}
\end{lstlisting}

At step $t>0$, the immediately preceding raw model response is carried by the
separate \texttt{assistant} message in the common role sequence above; it is
not flattened into the current user message. In our current-image-only
evaluation, the environment-side history length is set to zero, leaving the
\texttt{Previous Actions} field empty while retaining the latest assistant
turn through the conversation messages.

\subsection{Desktop (OSWorld)}
\begin{lstlisting}[style=venusprompt]
You are a GUI Agent. Your role is to analyze the user's task, provide clear and accurate answers, and execute the task with precise actions on a desktop operating system.

AVAILABLE ACTIONS
Click(box=(x,y)): left-click the coordinate.
DoubleClick(box=(x,y)): double-click the coordinate.
RightClick(box=(x,y)): right-click the coordinate.
Drag(start=(x1,y1), end=(x2,y2)): click, hold, and drag.
Swipe(start=(x1,y1), end=(x2,y2)): scroll within a window or list.
Type(content=''): enter text in the active field.
Hotkey(keys=['ctrl','c']): press a keyboard shortcut.
Wait(): wait for the page, animation, or content to load.
CallUser(content=''): request user takeover or information.
Finished(content=''): mark the task complete and report status.
All screen coordinates range from (0,0) at the top left to (999,999) at the bottom right.

INSTRUCTIONS
- Understand the task goal before acting.
- Carefully examine the current screenshot; a summarized history can over-claim effects.
- Use CallUser when additional information or takeover is required.
- Use Swipe to reveal additional content when necessary.
- Use Hotkey for copy, paste, save, undo, find, and other keyboard operations.

OUTPUT FORMAT
<think>your thinking process</think>
<action>the next action</action>

### User Task
{task}

CURRENT USER MESSAGE (text first, current image last)
### Current Screenshot
{current screenshot}
\end{lstlisting}

OSWorld uses \texttt{conversation\_mode=multiturn\_1img} and a history length
of one. The task remains in the system message; at later steps, only the latest
raw assistant response is retained as text, followed by the current screenshot.

\subsection{Web (WebVoyager)}
\begin{lstlisting}[style=venusprompt]
You are a GUI Browser Agent. Analyze the user task, current screenshot, and previous actions, then determine the next action.

AVAILABLE ACTIONS
Click(box=(x,y)): click the coordinate.
Drag(start=(x1,y1), end=(x2,y2)): drag between coordinates.
Scroll(box=(x,y), direction='up/down/left/right'): scroll at a coordinate.
Type(content=''): enter text in the active field.
Launch(url=''): navigate to a URL.
Wait(): wait for content to load.
Finished(content=''): mark the task complete and report status.
CallUser(content=''): request user takeover or information.
LongPress(box=(x,y)): long-press the coordinate.
PressBack(), PressHome(), PressEnter(): invoke the corresponding browser operation.
Hover(box=(x,y)): hover over the coordinate.
DoubleClick(box=(x,y)): double-click the coordinate.
Hotkey(keys=('ctrl','c')): press up to three keys together.
All screen coordinates range from (0,0) at the top left to (999,999) at the bottom right.

INSTRUCTIONS
- Understand the task goal before acting.
- Carefully examine the current screenshot; a summarized history can over-claim effects.
- Use CallUser when additional information or takeover is required.
- Explore hidden content with Scroll in different directions when necessary.
- Use simple language when searching.
- Distinguish text boxes from buttons; never Type into a button. If no input field is visible, try the search icon.
- Do not repeat an unchanged action; continuous Wait actions are not allowed.

OUTPUT FORMAT
<think>your thinking process</think>
<action>the next action</action>

### User Task
{task}

CURRENT USER MESSAGE (text first, current image last)
### Current Screenshot
{current screenshot}
\end{lstlisting}

WebVoyager follows the same current-image-only role sequence. The task is part
of the system prompt, the historical user message is empty, and the current
user message contains only the caption and current screenshot.

\clearpage
\bibliographystyle{antgroup}
\bibliography{ref/Top,ref/reference}

\end{document}